%% file: document.tex
\documentclass{article}

\include{shortcuts}
\usepackage[final]{nips_2017}

\usepackage[utf8]{inputenc} 
\usepackage[T1]{fontenc}    
\usepackage{hyperref}       
\usepackage{url}            
\usepackage{booktabs}       
\usepackage{amsfonts}       
\usepackage{nicefrac}       
\usepackage{microtype}      
\usepackage{courier}
\usepackage{graphicx}
\graphicspath{ {/} }
\usepackage{amsmath}
\usepackage{amssymb}
\usepackage{multirow}
\usepackage{graphicx}
\usepackage{subcaption}
\usepackage{courier}
\usepackage{bm}

\makeatletter
\renewcommand{\@noticestring}{}
\makeatother

\newcommand{\good}{\texttt{GOOD}}
\newcommand{\tip}{\texttt{TIP}}

\title{Generating Diverse and Informative Natural Language Fashion Feedback}

\author{
Gil~Sadeh\\
Amazon Lab126 \\
{\scriptsize \texttt{gilsadeh@amazon.com} }\\
\And
Lior~Fritz\\
Amazon Lab126 \\
{\scriptsize \texttt{liorf@amazon.com}} \\
\And
Gabi~Shalev\\
Amazon Lab126 \\
{\scriptsize \texttt{shalevg@amazon.com}} \\
\And
Eduard~Oks\\
Amazon Lab126 \\
{\scriptsize \texttt{oksed@amazon.com}} \\
}

\begin{document}

\maketitle

\begin{abstract}
	
	Recent advances in multi-modal vision and language tasks enable a new set of applications. 
	In this paper, we consider the task of generating natural language fashion feedback on outfit images. We collect a unique dataset, which contains outfit images and corresponding positive and constructive fashion feedback.
	We treat each feedback type separately, and train deep generative encoder-decoder models with visual attention, similar to the standard image captioning pipeline.
	Following this approach, the generated sentences tend to be too general and non-informative. We propose
	an alternative decoding technique based on the Maximum Mutual Information objective function, which leads to more diverse and detailed responses.
	We evaluate our model with common language metrics, and also show human evaluation results.
	This technology is applied within the ``Alexa, how do I look?'' feature, publicly available in Echo Look devices.

\end{abstract}

\section{Introduction}

Tasks combining image and language understanding, such as image captioning and visual question answering, continue to inspire research that combines computer vision and natural language processing (NLP). These are challenging tasks that open a rich set of interactive applications.

The task of image captioning, i.e. generating natural language image descriptions, has been studied thoroughly in recent years, especially following the release of the MS-COCO captioning challenge and dataset~\cite{lin2014microsoft, chen2015microsoft}. The most successful approach is based on training deep, end-to-end, encoder-decoder models, which yield impressive results~\cite{vinyals2015show,karpathy2015deep}. These models are comprised of a convolutional neural network (CNN) image encoder, and a recurrent neural network (RNN) decoder, and are trained to optimize the maximum likelihood (ML) objective function.

An additional extension of captioning models is the visual attention mechanism, which allows focusing on different image regions in each word prediction step, and leads to improved performance and generalization capabilities~\cite{xu2015show,lu2016knowing,rennie2016self,anderson2017bottom}. Recently,~\cite{anderson2017bottom} achieved state-of-the-art performance by combining bottom-up and top-down attention mechanisms, which enable the attention to be calculated at the level of objects and other salient image regions.

Image captioning algorithms are usually evaluated with several NLP metrics, such as BLEU~\cite{papineni2002bleu} or CIDEr~\cite{vedantam2015cider}, which are based on precision and recall of matching n-grams. Recent works used the REINFORCE algorithm~\cite{williams1992simple} to directly optimize these non-differentiable metrics and managed to reach state-of-the-art performance~\cite{ranzato2015sequence,rennie2016self, liu2016improved}.

Despite the substantial progress in recent years, sentences produced by existing image captioning methods are still often overly rigid and lacking in variability.
In~\cite{dai2017towards, shetty2017speaking} an alternative GAN-based training method was proposed to generate more natural and diverse image descriptions.

\newpage
In this paper, we address the task of generating fashion feedback about outfits. We collect a unique dataset for this task, containing outfit images and corresponding fashion feedback about what is good in the outfit and how to improve it. We treat each feedback type separately, and train generative models following the standard image captioning pipeline as described in Section~\ref{sec:training}.

Fashion feedback often relates to complex, subtle and abstract concepts (like fit, trend and color combination). Moreover, fashion recommendation feedback is often not directly visually grounded in the image. Thus, it is not trivial to expect the standard image captioning pipeline to generalize well to this task.

Unlike the standard image captioning task, in our case some very general responses coincide with many images (e.g., ``Your outfit fits you well''). Therefore, the standard pipeline leads to non-informative responses with low variability.
A similar issue is addressed in~\cite{lin2018netizen}, where a topic modeling based solution is proposed in order to increase the diversity of generated fashion comments.
We propose to utilize a decoding method based on the Maximum Mutual Information (MMI) objective, as suggested in~\cite{li2015diversity}, which incorporates an auxiliary language model that minimizes the effect of the sentence prior. This method yields considerable improvements in the diversity and specificity of the generated sentences.

This technology has a significant role in the ``Alexa, how do I look?'' feature, currently available in Echo Look devices. This feature allows customers to stand in front of this camera-enabled device, and ask for fashion feedback on their outfit. The device takes a picture, processes it and returns a vocal response.
We have found that improving the diversity and specificity of the responses leads to a better customer experience.

\section{Dataset}
We have targeted two general types of fashion feedback, which describe what is good in the outfit, and tips for improvement. We refer to these feedback types as \good{} and \tip{}, respectively. Figure~\ref{fig:examples} shows examples of generated sentences. With the help of a Fashion Specialists team, we have collected approximately 170K sentences, for each feedback type, on approximately 60K outfit images. A held-out set of 500 images, collected densely with approximately 15 captions per image, is used as an evaluation set. The annotation team was instructed to write sentences that are detailed and concise.

\subsection{Data Challenges}
\label{sec:data-challenges}
One of the main challenges in our task stems from its subjectivity. The collected sentences may often reflect the annotators personal styling taste and preferences. Moreover, the ambiguous nature of the task may lead to completely different outfit feedback, where one sentence may mention a certain fashion aspect or garment, and the others will refer to totally different aspects. This causes severe evaluation difficulties. Unlike traditional machine translation and image captioning tasks, our set of possible correct responses is much larger with a reduced tendency of using similar n-grams. Thus, most of the standard evaluation metrics, which rely on precision and recall of matching n-grams, do not reflect the performance of our models. And so, we rely heavily on qualitative tests and human evaluations. 

\begin{figure}
	\includegraphics[width=\textwidth]{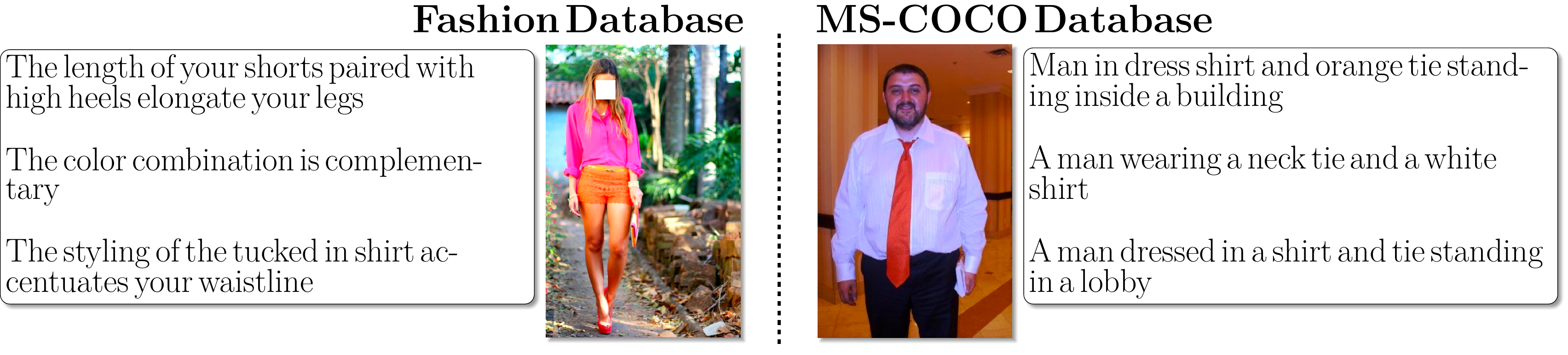}
	\centering
	\vspace{-15pt}
	\caption{Comparison between our data and MS-COCO data. In standard image captioning datasets, there are some valid and similar sentences per image. In our data, there are many valid and diverse sentences per image.}
	\label{fig:data_compare}
	\vspace{-10pt}
\end{figure}

Another challenge we face is the fact that the visual differences between images are very subtle. In the MS-COCO dataset~\cite{lin2014microsoft, chen2015microsoft}, image descriptions refer mostly to dominant objects and the relationships between them. In our case, all images include a human with more or less the same set of garments, so that the visual differences are much more fine-grained. Moreover, our text often refers to abstract fashion concepts. This requires both object recognition capabilities and fashion understanding. Figure~\ref{fig:data_compare} demonstrates the differences between our dataset and the MS-COCO captioning dataset.

Finally, a challenging aspect in our task is avoiding model degradation to generating very general responses, such as ``Your outfit fits you well''. These responses could be valid for a large set of images, yet we aim at generating diverse and specific responses in order to provide informative feedback. A solution to this problem is proposed in Section~\ref{sec:decoding_mmi}.

\section{Training Method}
\label{sec:training}
We train two separate models for each of the feedback types, \good{} and \tip{}. We adopt the standard image captioning training procedure for this task, which is typically done with an encoder-decoder mechanism. The encoder, which commonly consists of a convolutional neural network (CNN), transforms the image into visual features, and, in turn, a recurrent neural network (RNN) decodes these features in a generative process into a sequence of words (i.e. sentence).

The standard training procedure is termed \emph{Teacher Forcing}~\cite{williams1989learning}. The RNN receives the sub-sequence of previous ground-truth words as an input, and is trained to predict the next ground-truth word. Consider a database of images and corresponding sentences $ \{I_n, \bs_n\}  $ where $ I_n $ is an image, and $ \bs_n = (w_{n, 0}, \ldots, w_{n,T_n}) $ is a sequence of $ T_n $ words (i.e. sentence). The model parameters, $ \theta $, are trained to optimize the maximum likelihood (ML) objective function, $ \sum_n \log p(\bs_n \given I_n ; \theta) $, where

\begin{equation}
\log p(\bs \given I; \theta) = \sum_{t=1}^{T} \log p \left( w_{t} \given w_{0}, \ldots, w_{t-1}, I ; \theta \right) \,.
\end{equation}
The first and final words, $w_{0}$ and $w_{T}$, are special \emph{begin of sentence} and \emph{end of sentence} tokens.

We use a ResNet-18~\cite{he2016deep} CNN image encoder, pretrained on the ImageNet~\cite{deng2009imagenet}. 
Using a human detector (based on an SSD~\cite{liu2016ssd}, and trained on a separate dataset) we extract a bounding box of a human, resize it to a fixed size of $ 224 \times 448 $ and insert it as the image to our CNN. We extract the output of the last layer of spatial features, each of size 512, from the ResNet model. The CNN weights are fixed during the initial training steps, and after some epochs we propagate the gradients through them. Since the \tip{} feedback is often less visually grounded, we initialize the \tip{} model ResNet weights with the trained weights from the \good{} model.

\begin{figure}[h!]
	\includegraphics[width=0.6\textwidth]{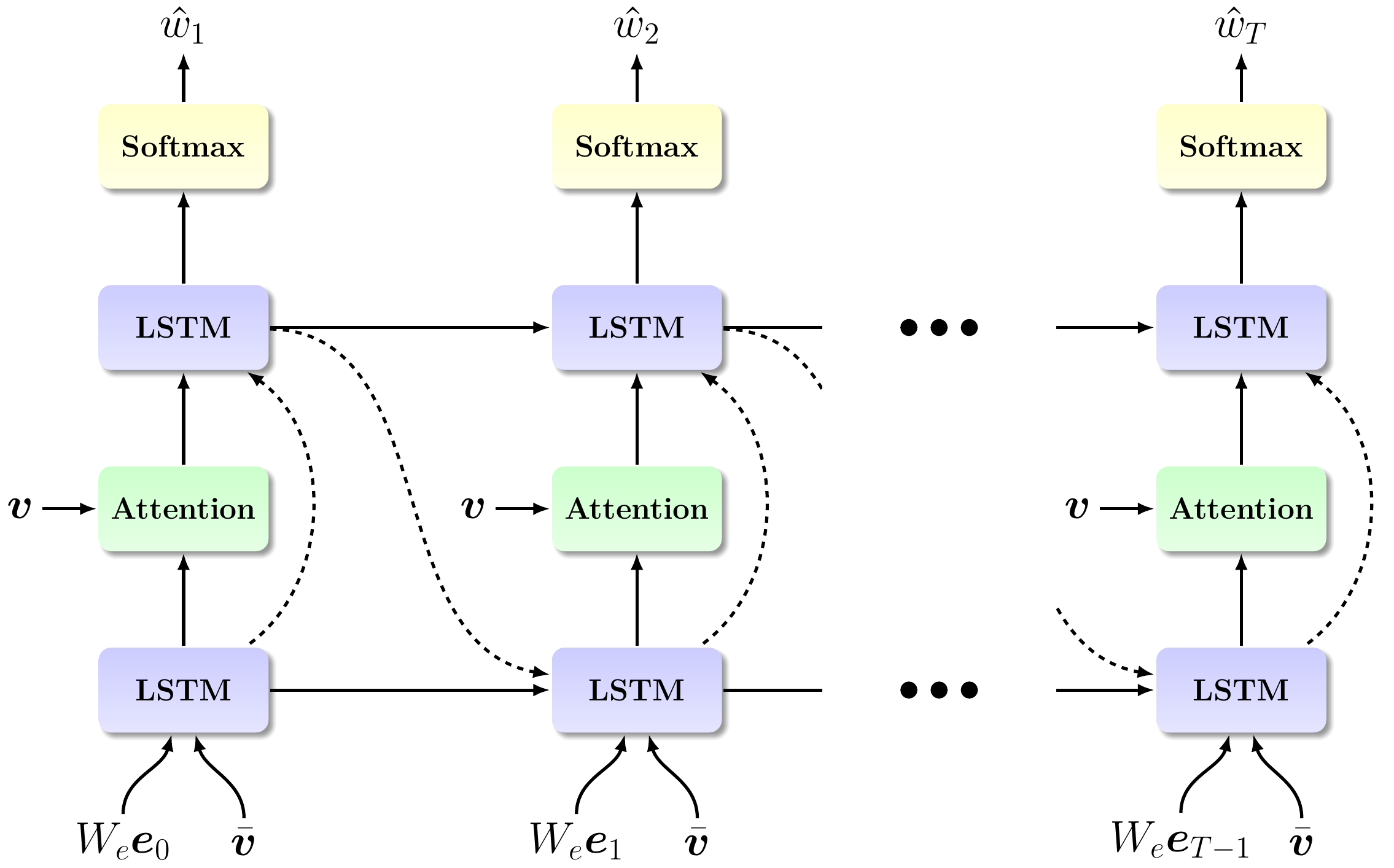}
	\centering
	\caption{The top-down architecture~\cite{anderson2017bottom} we use as our decoder. The first LSTM layer receives the averaged image features (over the spatial axes), $\bar {\bm v}$, and the embedded ground-truth words. It also receives the hidden state of the second LSTM layer as an input. $ W_e $ embeds the one-hot word vectors,  $ \{\bm e_t \}$, into vectors of size 512. $ \bm v $ is the spatial image features. The second LSTM layer receives both the attention mechanism output and the first layer output as inputs. The output of the second LSTM layer is followed by a softmax layer, and $ \{\hat w_t \}$ are the predicted words.}
	\label{fig:topdown}
	\vspace{-10pt}
\end{figure}

We use a RNN decoder module which is based on the top-down attention architecture as described in~\cite{anderson2017bottom}. 
The module consists of two long short-term memory (LSTM)~\cite{hochreiter1997long} layers. 
The first layer, termed the top-down attention LSTM, handles the attention mechanism. 
The second layer, termed the language LSTM, produces the prediction of the next word. Figure~\ref{fig:topdown} depicts the general structure of the decoder architecture.

The feed-forward procedure works as follows. Given an image $ I $ of a detected human, the encoder encodes the image into $ 7 \times 14 $ features of size 512. These features are fed through a fully connected layer with ReLU activation.
The parameters of this layer are trained jointly with the RNN decoder, unlike the parameters of the CNN, which are only trained after some epochs. Let $ \bm v $ be the output of these layers, which is of size $ 7 \times 14 $ in the spatial axes, and 512 in the features axis. At each step, $ t $, the input to the first LSTM layer is given by $\bm x_t^1 = [ \bm h_{t-1}^2, \bar {\bm v}, W_e \bm e_t]$,
where $ \bm h_{t-1}^2 $ is the output of the second LSTM layer at the previous step, $ \bar {\bm v} $ is the averaged image features (over the spatial axes),  $\bm e_t$ is the one-hot encoding of the word $ w_t $ and $ W_e $ is an embedding matrix. The output of the first LSTM layer, $ \bm h_t^1 $, along with the spatial image encoding, $ \bm v $, is used to calculate the attention score, for each $ i $ in the spatial axes,

\begin{equation}
\alpha_{i,t} = \bm u_\alpha^T \tanh \left(W_{v\alpha} \bm v_i + W_{h\alpha}\bm h_t^1 \right.) \,,
\end{equation}

where $ \bm u_\alpha $, $ W_{v \alpha} $ and $ W_{h \alpha} $ are learned parameters. The attention weights $\hat {\bm \alpha}_t$ are normalized with a softmax function,
\vspace{-1pt}
\begin{equation}
\hat{\bm \alpha}_t = {\rm softmax} \left(\bm \alpha_t\right) \,.
\end{equation}

Finally, the attended image encoding is calculated with a convex combination of all input features,
\begin{equation}
\hat{\bm v}_t = \sum_i \hat{\alpha}_{i,t} \bm v_i \,.
\end{equation}

The input to the language LSTM layer consists of the attended image feature, concatenated with the output of the attention LSTM, $\bm x_t^2= [\hat{\bm v}_t,\bm h_t^1]$.
The output of the second LSTM layer, $\bm h_t^2$, is used to calculate the conditional distribution over possible output words,
\begin{equation}
p\left(w_t | w_0, \ldots, w_{t-1}, I ; \theta \right.) = {\rm softmax} \left(W_p \bm h_t^2 + \bm b_p \right) \,,
\end{equation}
where $ W_p $ and $ \bm b_p $ are learned weights and biases, respectively.

\section{Sequence Generation}

At inference, ground truth labels are not given. The model predicts a sentence, word-by-word, given the image and its previous predictions. 
The standard inference objective is to find the sentence that maximizes the log likelihood of the sentence given the image,

\begin{equation}
\hat{\bs} = \argmax_{\bs} \log p(\bs | I) \,.
\end{equation}

Most image captioning works employ the beam search decoding technique in order to approximate this objective. 
In beam search~\cite{koehn2009statistical}, at each step, a set of $ k $ (beam width) most probable candidate sentences are considered based on the log likelihood of the sub-sequences and the next word candidates. 

Empirically, we have found that increasing the beam width tends to degrade the generated sentences in terms of diversity and specificity. 
Since general and vague sentences can be paired with many images, and may also occur more often in our training set, the generative model would tend to favor these sentences. This happens due to the definition of the decoding objective function, which is aimed at finding the sentence that is most probable given the image.

\subsection{Decoding with Maximum Mutual Information}
\label{sec:decoding_mmi}
In order to account for the issues posed by the standard decoding objective, we suggest to change it into the Maximum Mutual Information (MMI) objective function, as done in~\cite{li2015diversity}, where the task of conversation response generation is considered. It was found that the MMI decoding objective leads to more specific and diverse responses. The MMI decoding objective is defined by

\begin{align}
\hat{\bs} =  & \argmax_{\bs} \left\{\log  \frac{p(\bs , I)}{p(\bs)p(I)}\right\} =  \argmax_{\bs} \left\{ \log p(\bs \given I) -  \log p(\bs) \right\} =  \argmax_{\bs}  \log p(I \given \bs) \,.
\end{align}

This objective increases the specificity and diversity of the generated sentences, at the cost of allowing grammatically incorrect sentences.
Hence, a compromised solution is considered. As in~\cite{li2015diversity}, we use a weighted average of the two objectives 

\begin{align}
\hat{\bs} = \argmax_{\bs} \left\{(1-\beta) \log p(\bs \given I) +  \beta \log p(I \given \bs) \right\} = \argmax_{\bs} \left\{\log p(\bs \given I) -  \beta \log p(\bs) \right\} \,,
\end{align}
where $ \beta \in [0,1] $ is a hyperparameter,  $ \log p(\bs \given I) $ is given by our trained model and $ \log p(\bs)  $ is given by an auxiliary language model, as explained in Section~\ref{sec:LM}.

\subsection{Auxiliary Language Model}
\label{sec:LM}

In order to apply the MMI criterion, an auxiliary language model (LM) is required. We consider a simple RNN LM which is comprised of one LSTM layer. The model is also trained with \emph{Teacher Forcing}. The model parameters, $ \phi, $ are trained to optimize the ML objective function, $ \sum_n \log p(\bs_n ; \phi) $, where
\vspace{-4pt}
\begin{equation}
\log p(\bs; \phi) = \sum_{t=1}^{T} \log p \left( w_{t} \given w_{0}, \ldots, w_{t-1}; \phi \right) \,.
\end{equation}

This model can approximate a sentence likelihood, $p(\bs)$.

\subsection{Language Mistakes}

Captioning models are known to produce language mistakes in some cases~\cite{devlin2015language}. As discussed in \ref{sec:decoding_mmi}, the introduction of the MMI objective further increases the rate of language mistakes generated by our model. We have found that these mistakes mostly occur within \emph{noun phrases} in the sentences, while the other grammatical aspects of the sentences remain valid. A noun phrase is a noun with its preceding words which help define it.
We have defined the following set of rules, which detect most of these common language mistakes. For \good{} and \tip{}, a word cannot repeat in a noun phrase (``Add a black striped black jacket''). For \good{}, a noun cannot repeat (``Your leggings complement your black leggings''). For \tip{}, we consider these cases as valid (``Swap your black leggings for white leggings''), however complete noun phrases repetition is forbidden (``Swap your black leggings for black leggings'').

We perform \emph{noun phrase chunking} of the sentence with the Spacy toolkit~\footnote{\url{https://spacy.io/}}. Since the model generates a number of sentences during the beam search procedure, we can filter sentences which do not follow our set of rules.

\section{Experimental Results}
\label{sec:exp}

We train both captioning and language models for each of the question types, \good{} and  \tip{}.
The vocabulary is built for each question type separately, containing only words which appear more than 5 times in the training set.
All other words are encoded as a special \emph{unknown} token.
The sentences are also pre-processed by removing punctuation marks, except for commas, and changing all letters to lowercase. The resulting vocabulary sizes are 1,745 and 1,807 words for the \good{} and \tip{} models, respectively.

We train the captioning models using the ADAM~\cite{kingma2014adam} optimizer.
We use an effective batch size of 96, which is composed of 32 images with 3 sentences each.
We also employ dropout~\cite{srivastava2014dropout} with probability 0.5, both in the final fully connected layer and in the word embedding layer.
We train the network for some epochs (10-15) before unfreezing the CNN weights. The best model is picked according to its CIDEr-D score on the evaluation set.

\subsection{MMI Decoding Analysis}
\begin{figure}[h!]
	
	\begin{subfigure}[t]{\textwidth}
		
		\includegraphics[width=0.9\textwidth]{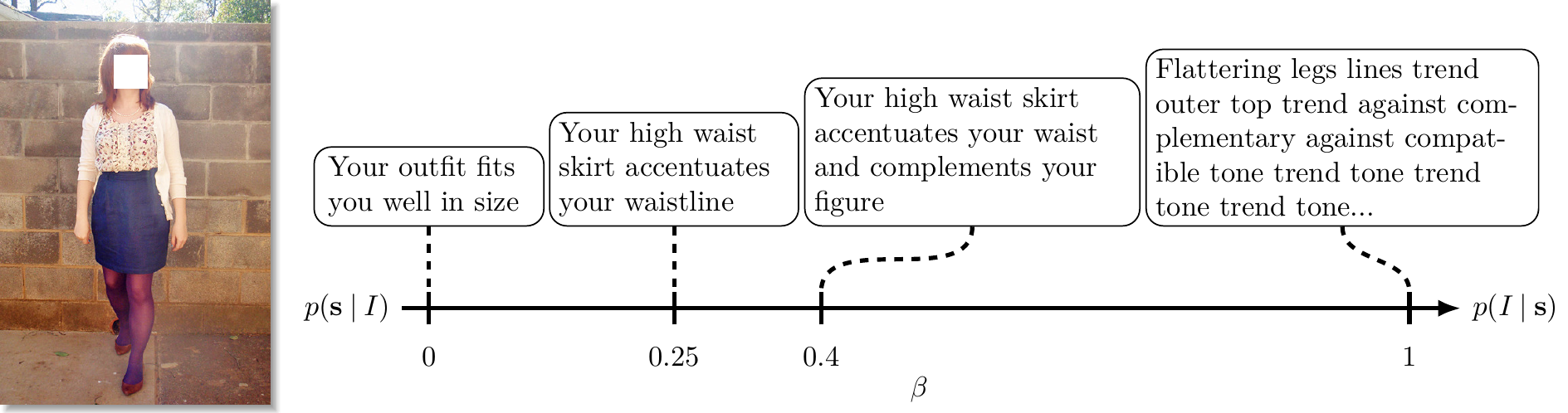}
		\centering
		\caption{}
		\label{fig:beta-example}
	\end{subfigure}
	
	\begin{subfigure}[t]{0.5\textwidth}
		\includegraphics[width=\textwidth]{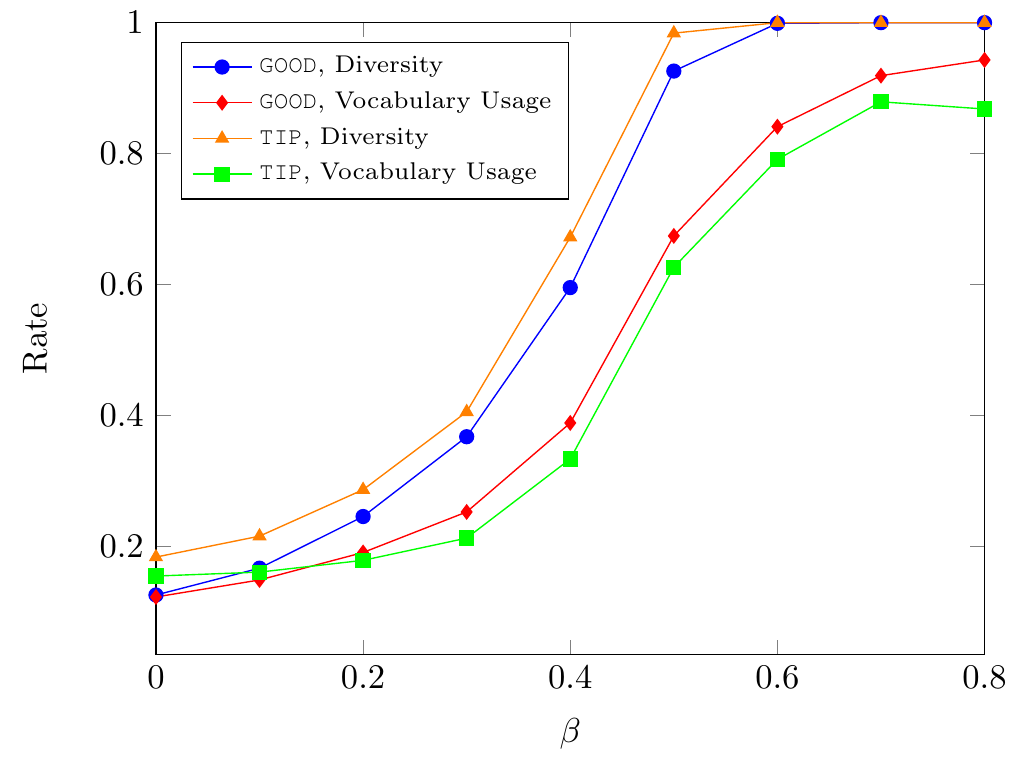}
		\centering
		\caption{}
		\label{fig:beta-div}
	\end{subfigure}
	~
	\begin{subfigure}[t]{0.5\textwidth}
		\includegraphics[width=\textwidth]{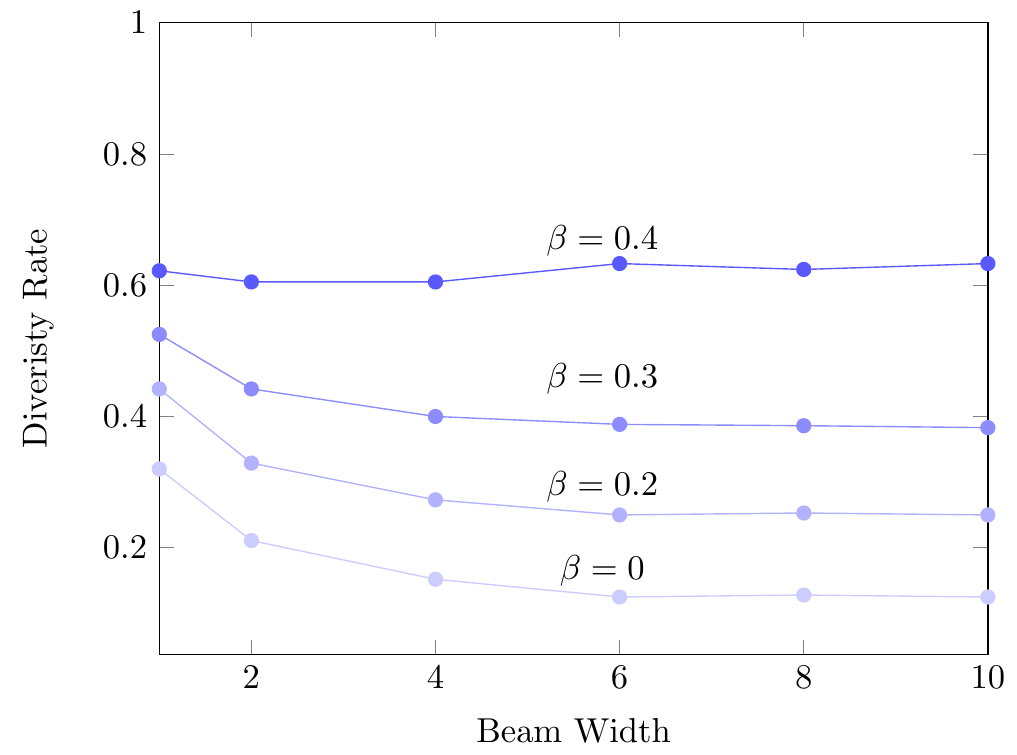}
		\centering
		\caption{}
		\label{fig:div-bs-beta}
	\end{subfigure}
	\caption{The effect of beam width and $ \beta $ on the diversity and vocabulary usage. (\subref{fig:beta-example}) demonstrates the qualitative effect of $\beta$ on generated sentences. When it is too low, the sentence is general and uninformative, while very high values may lead to grammatical mistakes. (\subref{fig:beta-div}) shows the quantitative effect of $\beta$ on diversity and vocabulary usage of generated sentences. Beam width is set to 10 in these experiments. (\subref{fig:div-bs-beta}) demonstrates the effect of beam width on the diversity, with different values of $ \beta $, on the \good{} model. A larger beam width provides a better approximation of the decoding objective, and produces more stabilized results. Yet, diversity degradation is apparent when the beam width is increased in standard decoding ($\beta=0$). This effect is diminished with larger $\beta$ values.}
	\label{fig:beta-exp}

\end{figure}

We experiment with MMI decoding using various values of $\beta$ in order to demonstrate its effect on the generated sentences.
We evaluate the specificity and diversity of a given model and decoding technique by measuring 2 metrics, computed on the corpus of generated responses on images from our evaluation set. 
The \emph{Diversity} metric measures the rate of unique generated sentences. The \emph{Vocabulary~Usage} metric captures the rate of words appearing at least once in the generated text corpus.
We demonstrate the qualitative and quantitative effects of $\beta$ in Figure~\ref{fig:beta-exp}.

\newpage
\begin{table}[h!]
	\begin{center}
		\bgroup
		\def\arraystretch{1.2}
		
		\caption{Evaluation results on standard captioning metrics, diversity (Div.) and vocabulary usage (Vocab.). In all metrics, higher is better. The FC model is the model without attention. FS stands for Fashion Specialists.}
		\label{table:metric-eval}
		\begin{tabular}{|l | l|c|c|c|c|c|c|}
			
			\hline
			\multicolumn{2}{|l|}{}    		  &BLEU4 &ROUGE-L &METEOR & CIDEr-D & Div. &Vocab. \\
			\hline
			\hline
			
			\multirow{4}{*}{\good{}} & FC model            &0.582  &0.671  &0.331   &0.421 &0.158  &0.061 \\
			\cline{2-8}
			& Top-Down         & 0.62    & 0.706 &0.364  &0.593 &0.43 &0.096 \\
			\cline{2-8}
			& Top-Down + MMI          & 0.604  & 0.694 & 0.365  &0.743 &0.848 &0.127 \\
			\cline{2-8}
			& FS performance       		  & 0.341  & 0.55 & 0.29  &0.438 &0.965&0.281\ \\
			\hline
			\hline
			\multirow{4}{*}{\tip{}} & FC model              & 0.46  &0.599  &0.293   &0.244 &0.096 &0.051 \\
			\cline{2-8}
			& Top-Down        &0.512& 0.631 & 0.317 &0.256 & 0.335 & 0.11 \\
			\cline{2-8}
			& Top-Down + MMI         & 0.518  & 0.623 & 0.317  &0.38 &0.737 &0.141 \\
			\cline{2-8}
			& FS performance       & 0.295  & 0.486 & 0.242  &0.184 &0.988&0.287\\
			
			\hline
		\end{tabular}
		\egroup
		
	\end{center} 
		\vspace{-15pt}
	
\end{table}

\subsection{Automatic Metrics Results}
Traditionally, captioning tasks are evaluated with NLP metrics, based mostly on n-gram matching, such as BLEU~\cite{papineni2002bleu}, ROUGE~\cite{lin2004rouge}, METEOR~\cite{lavie2005meteor} or CIDEr~\cite{vedantam2015cider}. 
We found that CIDEr is the only NLP metric which captures diversity and specificity. This is due to the fact that unlike other metrics, it rewards accurate generation of less frequent n-grams.

As a baseline, we compare our top-down architecture to a standard captioning model without an attention mechanism, in which a single LSTM layer receives the current word embedding at every step and predicts the following word. 
In this model, the image embedding is projected onto the word embedding subspace, and is inserted as input to the LSTM in the first step.
We also compare our MMI decoding technique with the standard decoding objective. 
In addition, we estimate the performance of Fashion Specialists by selecting a random ground truth sentence for each image, and evaluating it based on the remaining ground truth sentences.

\begin{figure}[h!]
	\vspace{20pt}
	\includegraphics[width=\textwidth]{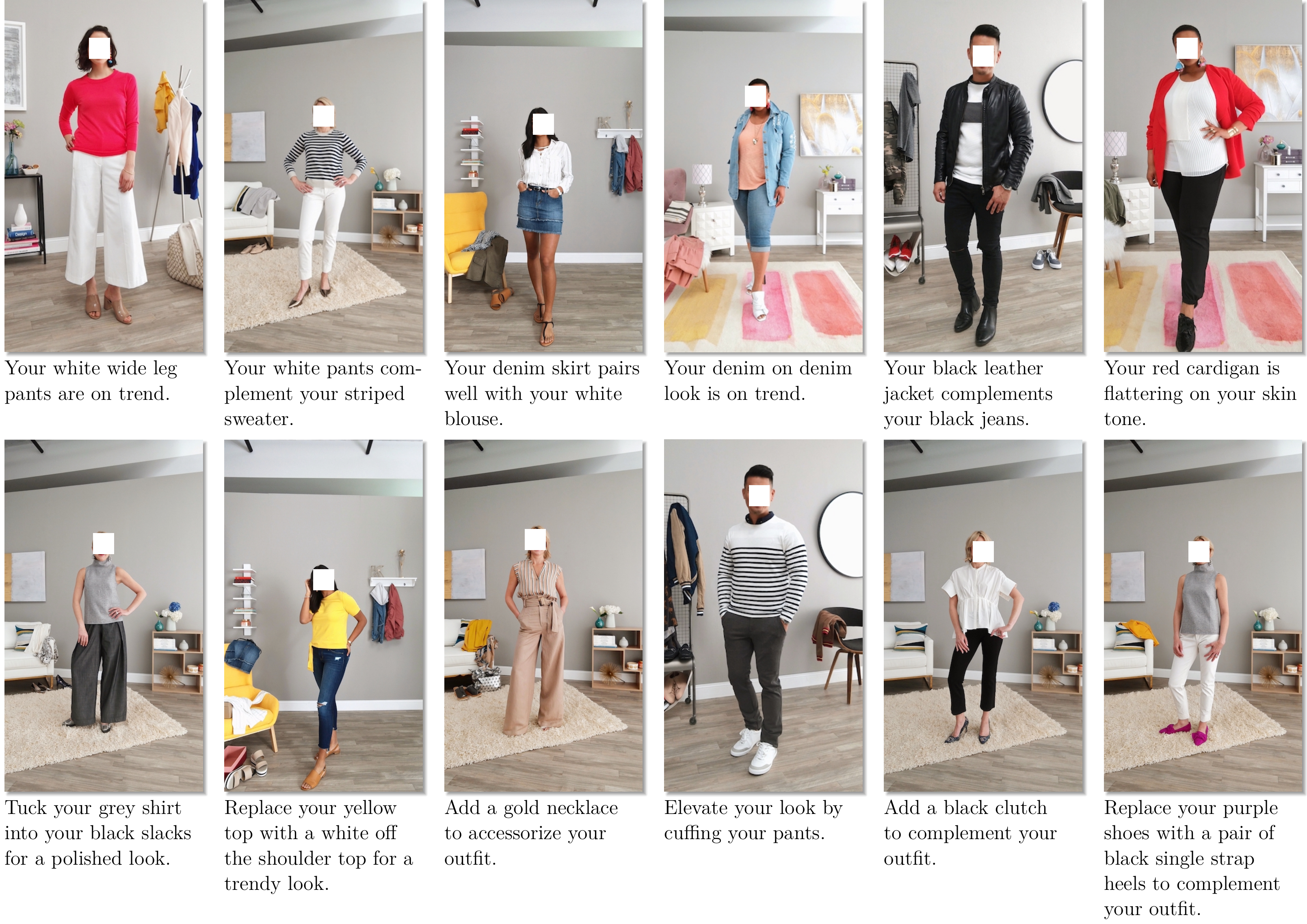}
	\centering
	\caption[Example Caption]{Examples of outfit images~\footnotemark  and corresponding generated sentences from the \good{} (top row) and \tip{} (bottom row) models.
}

	\label{fig:examples}
\end{figure}

In Table~\ref{table:metric-eval} we report the performance of our best model in comparison to the mentioned baselines, on the evaluation set. In our best model, we incorporate the top-down architecture alongside MMI decoding with $ \beta=0.4 $.
We set $ \beta=0 $ after 11 and 16 steps in the \good{} and \tip{} models, respectively. This decreases the rate of language mistakes, while not harming the diversity significantly, as also reported in~\cite{li2015diversity}.
It can be observed from the table, that incorporating attention improves all metrics, and using the MMI objective further improves the CIDEr, diversity and vocabulary usage metrics. It is also evident that our best model surpasses the Fashion Specialists performance in all standard NLP metrics. This shows the discrepancy of these metrics for our task. However, a gap remains in terms of diversity and vocabulary usage.
Figure~\ref{fig:examples} shows examples of generated responses.

\subsection{Human Evaluation}
We perform an extensive human evaluation study in order to get a better qualitative evaluation of the responses our algorithm produces.
In this study, we show Fashion Specialists images with sentences from the evaluation set, where the sentences are randomly selected either from the ground truth sentences or from the generated ones.
We perform 2 types of human evaluations. The \emph{Fashion} test is a fashionable variation of a Turing Test, where we ask whether the response was written by a Fashion Specialist on the given image. 
In addition, we wish to isolate the language aspect.
The \emph{Language} test is based on the sentence alone, without relating to the correspondence with the image. The evaluator is asked whether the response was written by a human, and is free of any language mistakes. Each annotation is performed 3 times by different Fashion Specialists.
The results are presented in Table~\ref{table:human-eval}. 

The overall performance of our models does not fall short considerably compared to the performance of the Fashion Specialists. Moreover, the results on ground truth responses in these tests can shed light on some of the problems in our dataset, which is not free of language and fashion mistakes. In the Fashion Turing Test, it can be observed that our generated responses managed to frequently fool the evaluators. Regarding the language aspect, the generated responses achieve comparable results to human performance.
\begin{table}[ht!]
	\begin{center}
		\caption{Human evaluation results. The numbers represent the rate of responses passing each test.}
		\vspace{5pt}
		\bgroup
		\def\arraystretch{1.2}
		
		\begin{tabular}{|l | l|c|c|}
			\hline
			\multicolumn{2}{|l|}{}    		  & Fashion & Language \\
			\hline
			\hline
			\multirow{2}{*}{\good{}} & Generated              & 0.83  & 0.89  \\
			\cline{2-4}
			& Ground Truth         & 0.88   & 0.86 \\
			\hline
			\hline
			\multirow{2}{*}{\tip{}} & Generated              & 0.84  & 0.93 \\
			\cline{2-4}
			& Ground Truth         & 0.90   & 0.88  \\
			\hline
		\end{tabular}
		\egroup
		
		\label{table:human-eval}
	\end{center}
\end{table}

\newpage
\section{Conclusions}
\label{sec:conclusion}

In this paper, we explored the task of fashion feedback generation in natural language. We trained end-to-end encoder-decoder models that generate sentences about what is good in the outfit, and how to improve it. We employed a decoding technique based on the MMI objective function in order to obtain diverse and specific generated responses.

These experiments show that this type of models can learn complex, subtle, abstract and non-directly visually grounded concepts. Results demonstrate that the performance of our models is comparable to Fashion Specialists in the measured metrics. The models are deployed within the ``Alexa, how do I look?'' feature, available in Echo Look devices.

\footnotetext{Example images belong to Amazon marketing, and are not part of our dataset.}
\newpage

\medskip

\small
\bibliographystyle{plain}

\bibliography{all_bib}


\end{document}

%% file: shortcuts.tex
\newcommand{\rem}[1]{}

\newcommand{\bre}{\begin{equation}}
\newcommand{\ere}{\end{equation}}

\newcommand{\ee}\]
\newcommand{\bra}{\begin{eqnarray}}
\newcommand{\era}{\end{eqnarray}}
\newcommand{\bfg}{\begin{figure}[hbtp]}
\newcommand{\efg}{\end{figure}}
\newcommand{\bver}{\begin{verbatim}}
\newcommand{\ever}{\end{verbatim}}
\newcommand{\bit}{\begin{itemize}}
\newcommand{\eit}{\end{itemize}}
\newcommand{\ben}{\begin{enumerate}}
\newcommand{\een}{\end{enumerate}}

\newcommand{\given}{\: | \:}

\newcommand{\bm}{{\bf m}}

\newcommand{\baa}{\begin{eqnarray*}}
\newcommand{\eaa}{\end{eqnarray*}}

\newcommand{\bs}{{\bf s}}

\def\argmax{\mathop{\rm argmax}}

\newfont{\boldlarge}{msbm10 scaled 1100}

\newcommand{\comment}[1]{}

\newlength{\tmpbigbar}
